\documentclass{llncs}
\usepackage{mathtools}
\usepackage{amssymb}

\usepackage{paralist}

\usepackage[numbers]{natbib}
\newcommand{\vc}[1]{\mathbf{#1}}%
\DeclareMathOperator*{\argmin}{arg\,min}

\DeclareMathOperator*{\Tr}{Tr}

\usepackage{arydshln}

\usepackage{hyperref}

\usepackage{caption}
\usepackage{subcaption}
\begin{document}

\title{Forecasting and Granger Modelling with Non-linear Dynamical Dependencies}
\titlerunning{Forecasting with Non-linear Dynamical Dependencies}

\author{Magda Gregorov\'a\inst{1,2} \and
        Alexandros Kalousis\inst{1,2} \and
        St\'ephane Marchand-Maillet\inst{2}
}

\institute{Geneva School of Business Administration, HES-SO University of Applied Sciences of Western Switzerland \and University of Geneva, Switzerland}

\maketitle

\begin{abstract}
Traditional linear methods for forecasting multivariate time series are not able to satisfactorily model the non-linear dependencies that may exist in non-Gaussian series.
We build on the theory of learning vector-valued functions in the reproducing kernel Hilbert space and develop a method for learning prediction functions that accommodate such non-linearities.
The method not only learns the predictive function but also the matrix-valued kernel underlying the function search space directly from the data.
Our approach is based on learning multiple matrix-valued kernels, each of those composed of a set of input kernels and a set of output kernels learned in the cone of positive semi-definite matrices.
In addition to superior predictive performance in the presence of strong non-linearities, our method also recovers the hidden dynamic relationships between the series and thus is a new alternative to existing graphical Granger techniques.
\end{abstract}

\section{Introduction}
Traditional methods for forecasting stationary multivariate time series
from their own past
are derived from the classical linear ARMA modelling.
In these, the prediction of the next point in the future of the series is constructed as a linear function of the past observations.
The use of linear functions as the predictors is in part based on the Wold representation theorem (e.g. \cite{Brockwell2006})
and in part, probably more importantly, on the fact that the linear predictor is the best predictor (in the mean-square-error sense) in case the time series is Gaussian.

The Gaussian assumption is therefore often adopted in the analysis of time series to justify the simple linear modelling.
However, it is indeed a simplifying assumption since for non-Gaussian series the best predictor may 
very well be a non-linear function of the past observations. 
A number of parametric non-linear models has been proposed in the literature, each adapted to capture specific sources of non-linearity (for example multiple forms of regime-switching models, e.g. \cite{Turkman2014}).

In this paper we adopt an approach that does not rely on such prior assumptions for the function form.
We propose to learn the predictor as a general vector-valued function $\vc{f}$ that takes as input the past observations of the multivariate series and outputs the forecast of the unknown next value (vector).

We have two principal requirements on the function $\vc{f}$.
The first is the standard prediction accuracy requirement.
That is, the function $\vc{f}$ shall be such that we can expect its outputs to be close (in the squared error sense) to the true future observations of the process.
The second requirement is that the function $\vc{f}$ shall 
have a structure that will
enable 
the analysis of the relationships amongst the subprocesses of the multivariate series.
Namely, we wish to understand how parts of the series help in forecasting other parts of the multivariate series, a concept known in the time-series literature as graphical Granger modelling \cite{Granger1969,Eichler2012}.

To learn such a function $\vc{f}$ we employ the framework of regularised learning of vector-valued functions in the reproducing kernel Hilbert space (RKHS) \cite{Micchelli2005b}.
Learning methods based on the RKHS theory have previously been considered for time series modelling (e.g. \cite{Franz2006,Sindhwani2014,Lim2014}).
Though, as Pillonetto et al. note in their survey \cite{Pillonetto2014}, their adoption for the dynamical system analysis is not a commonplace.

A critical step in kernel-based methods for learning vector-valued functions is the specification of the operator-valued kernel that exploits well the relationships between the inputs and the outputs.
A convenient and well-studied class of operator-valued kernels  (e.g. in \cite{Caponnetto2008,Dinuzzo2011a,Jawanpuria2015}) are those decomposable into a product of a scalar kernel on the input space (input kernel) and a linear operator on the output space (output kernel).

The kernel uniquely determines the function space within which the function $\vc{f}$ is learned.
It thus has significant influence on both our objectives described above.
Instead of having to choose the input and the output kernels a priori, we introduce a method for learning the input and output kernels from the data together with learning the vector-valued function $\vc{f}$.

Our method combines in a novel way the multiple-kernel learning (MKL) approach \cite{Lanckriet2004} with learning the output kernels within the space of positive semidefinite linear operators on the output space \cite{Jawanpuria2015}.
MKL methods for operator-valued kernels have recently been developed in \cite{Kadri2012} and \cite{Sindhwani2014}.
The first learns a convex combination of a set of operator-valued kernels fixed in advance, the second combines a fixed set of input kernels with a single learned output kernel.
To the best of our knowledge, ours is the first method in which the operator-valued kernel is learned by combining a set of input kernels with a set of multiple learned output kernels.

In accordance with our second objective stated above, we impose specific structural constraints on the function search space so that the learned function supports the graphical Granger analysis.
We achieve this by working with matrix-valued kernels operating over input partitions restricted to single input scalar series
(similar input partitioning has recently been used in \cite{Sindhwani2014}).

We impose diagonal structure on the output kernels to control the model complexity.
Though this has a cost in the inability to model contemporaneous relationships, it addresses the strong over-parametrisation in a principled manner.
It also greatly simplifies the final structure of the problem, which, in result, suitably decomposes into a set of smaller independent problems solvable in parallel.

We develop two forms of sparsity-promoting regularisation approaches for learning the output kernels.
These are based on the $\ell_1$ and $\ell_1/\ell_2$ norms respectively and are motivated by the search for Granger-causality relationships.
As to our knowledge, the latter has not been previously used in the context of MKL.

Finally, we confirm on experiments the benefits our methods can bring to forecasting non-Gaussian series in terms of improved predictive accuracy and the ability to recover hidden dynamic dependency structure within the time series systems.
This makes them valid alternatives to the state-of-the-art graphical Granger techniques.

\paragraph{Notation}
We use bold upper case and lower case letters for matrices and vectors respectively, and the plain letters with subscripts for their elements.
For any matrix or vector the superscript $^T$ denotes its transpose.
Vectors are by convention column-wise so that $\vc{x} = (x_1, \ldots, x_n)^T$ is the $n$-dimensional vector $\vc{x}$.
$\mathbb{R}, \mathbb{R}^n, \mathbb{R}^{m \times n}$ are the sets of real scalars, $n$-dimensional vectors, and $m \times n$ dimensional matrices.
$\mathbb{R}^{m \times n}_+$ is the set of non-negative matrices, $\mathbb{S}^{m}_+$ the set of positive semi-definite $m \times m$ matrices and $\mathbb{D}^{m}_+$ the set of non-negative diagonal matrices.
$\mathbb{N}_m$ is the set of positive integers $\{1,\ldots,m\}$.
For any vectors $\vc{x, y} \in \mathbb{R}^n$, $\langle \vc{x},\vc{y} \rangle, ||\vc{x}||_1, ||\vc{x}||_2$ are the standard inner product, $\ell_1$ and $\ell_2$ norms in the real Hilbert spaces.
For any square matrix $\vc{A}$, $\Tr(\vc{A})$ denotes the trace. For any two matrices $\vc{A, B} \in \mathbb{R}^{m \times n}$, $\langle \vc{A},\vc{B} \rangle_F := \Tr(\vc{A}^T \vc{B})$ is the Frobenius inner product and $||\vc{A}||_F := \sqrt{\langle \vc{A},\vc{A} \rangle_F} $ the Frobenius norm.
$\langle .,. \rangle_{\mathcal{F}}$ and $||.||_{\mathcal{F}}$ are the inner product and norm in the Hilbert space $\mathcal{F}$.

\section{Problem Formulation}\label{sec:ProblemFormulation}

Given a realisation of a discrete stationary multivariate time series process $\big\{ \vc{y}_t \in \mathcal{Y} \subseteq \mathbb{R}^m: t \in \mathbb{N}_n\}$,
our goal is to learn a vector-valued function $\vc{f}: \mathcal{Y}^p \to \mathcal{Y}$ 
that takes as input the $p$ past observations of the process and predicts its future vector value (one step ahead).
The function $\vc{f}$ shall be such that 
\begin{compactenum}
%\begin{enumerate}
\item we can expect the prediction to be near (in the Euclidean distance sense) the unobserved future value
\item its structure allows to analyse if parts (subprocesses) of the series are useful for forecasting other subprocesses within the series or if some subprocesses can be forecast independently of the rest; in short, it allows Granger-causality analysis \cite{Granger1969,Eichler2012}.
%\end{enumerate}
\end{compactenum}

For notational simplicity, from now on we indicate the output of the function $\vc{f}$ as $\vc{y} \in \mathcal{Y} \subseteq \mathbb{R}^{m}$ and the input as $\vc{x} \in \mathcal{X} \subseteq \mathbb{R}^{mp}$ (bearing in mind that $\mathcal{X} = \mathcal{Y}^p$ is in fact the $p$-th order Cartesian product of $\mathcal{Y}$ and that the inputs $\vc{x}$ and outputs $\vc{y}$ are the past and future observations of the same $m$-dimensional series).
We also align the time indexes so that our data sample consists of input-output data pairs $\big\{ (\vc{y}_t,\vc{x}_t) : t \in \mathbb{N}_n \big\}$.

Following the standard function learning theory, we will learn $\vc{f} \in \mathcal{F}$ by minimising the regularised empirical squared-error risk (with a regularization parameter $\lambda > 0$)
\begin{eqnarray}\label{eq:GenMin}
& \vc{\widehat{f}} = \argmin_{\vc{f} \in \mathcal{F}} \, \mathit{R}(\vc{f}) & \nonumber \\
& \mathit{R}(\vc{f}) := \sum_{t=1}^T ||\vc{y}_t - \vc{f}(\vc{x}_t)||_2^2 
+ \lambda \, ||\vc{f}||_{\mathcal{F}}^2 & \enspace .
\end{eqnarray}
Here $\mathcal{F}$ is the reproducing kernel Hilbert space (RKHS) of $\mathbb{R}^m$-valued functions endowed with the norm $||.||_{\mathcal{F}}$ and the inner product $\langle ., . \rangle_\mathcal{F}$.
The RKHS is uniquely associated with a symmetric positive-semidefinite matrix-valued kernel $\vc{H}: \mathcal{X} \times \mathcal{X} \to \mathbb{R}^{m \times m}$ with the reproducing property
\begin{equation*}
\langle \vc{y}, \vc{g}(\vc{x}) \rangle = \langle \vc{H}_\vc{x} \, \vc{y}, \vc{g}  \rangle_\mathcal{F} \quad \forall (\vc{y,x,g}) \in (\mathcal{Y},\mathcal{X},\mathcal{F}) \enspace ,
\end{equation*}
where the map $\vc{H}_\vc{x} : \mathcal{X} \to \mathbb{R}^{m \times m}$ is the kernel section of $\vc{H}$ centred at $\vc{x}$ such that $\vc{H}_{\vc{x}_i}(\vc{x}_j) = \vc{H}(\vc{x}_i,\vc{x}_j)$ for all $(\vc{x}_i,\vc{x}_j) \in (\mathcal{X},\mathcal{X})$.
From the classical result in \cite{Micchelli2005b},
the unique solution $\vc{\widehat{f}}$ of the variational problem \eqref{eq:GenMin} admits a finite dimensional representation
\begin{equation}\label{eq:ReprTheor}
\vc{\widehat{f}} = \sum_{t=1}^T \vc{H}_{\vc{x}_t} \, \vc{c}_t \enspace ,
\end{equation}
where the coefficients $\vc{c}_t \in \mathcal{Y}$ are the solutions of the system of linear equations
\begin{equation}
\sum_{t=1}^T \big( \vc{H}(\vc{x}_s, \vc{x}_t) + \lambda \delta_{st}   \big) \, \vc{c}_t = \vc{y}_s, \quad \forall s \in \mathbb{N}_n \enspace ,
\end{equation}
where $\delta_{st} = 1$ if $s = t$ and is zero otherwise.

\subsection{Granger-causality Analysis}\label{sec:Granger}

To study the dynamical relationships in time series processes, Granger \cite{Granger1969} proposed a practical definition of causality 
based on the accuracy of least-squares predictor functions. 
In brief, for two time series processes $\vc{y}$ and $\vc{z}$, $\vc{y}$ is said to Granger-cause $\vc{z}$ ($\vc{y} \to \vc{z}$) if given all the other relevant information we can predict the future of $\vc{z}$ better (in the mean-square-error sense) using the history of $\vc{y}$ than without it.

Though the concept seems rather straightforward, there are (at least) three points worth considering.
First, the notion is purely technical based on the predictive accuracy of functions with differing input sets; it does \emph{not} seek to understand the underlying forces driving the relationships.
Second, in practice the conditioning set of information needs to be reduced to all the \emph{available} information instead of all the \emph{relevant} information.
Third, it only considers relationships between pairs of (sub-)processes and not the interactions amongst a set of series.

Eichler \cite{Eichler2012} extended the concept to multivariate analysis through graphical models.
The discussion in the paper 
focuses on the notion of Granger non-causality rather than causality and 
describes the specific Markov properties (conditional non-causality) encoded in the graphs of Granger-causal relationships.
In this sense, the absence of a variable in a set of inputs is more informative of the Granger (non-)causality than its presence.
In result, graphical Granger methods are typically based on (structured) sparse modelling \cite{Bahadori2013}.

\section{Function Space and Kernel Specification}\label{sec:FuncSpace}

The function space $\mathcal{F}$ within which $\vc{f}$ is learned is fully determined by the reproducing kernel $\vc{H}$.
Its specification
is therefore critical for achieving the two objectives for the function $\vc{f}$ defined in Sect. \ref{sec:ProblemFormulation}.
We focus on the class of matrix-valued kernels decomposable into the product of input kernels, capturing the similarities in the inputs, and output kernels, encoding the relationships between the outputs.

To analyse the dynamical dependencies between the series, we need to be able to discern within the inputs of the learned function $\vc{f}$ the individual scalar series.
Therefore we partition the elements of the input vectors according to the source scalar time series.
In result, instead of a single kernel operating over the full vectors, we work with multiple partition-kernels, each of them operating over a single input series.
We further propose to learn the partition-kernels by combining the MKL techniques with output kernel learning within the cone of positive semi-definite matrices.

More formally, the kernel we propose to use is constructed as a sum of kernels $\vc{H} = \sum_j^m \vc{H}^{(j)}$, where $m$ is the number of the individual scalar-valued series in the multivariate process (dimensionality of the output space $\mathcal{Y}$).
Each $\vc{H}^{(j)} : \mathcal{X}^{(j)} \times \mathcal{X}^{(j)} \to \mathbb{R}^{m \times m}$ 
is a matrix-valued kernel that determines its own RKHS of vector-valued functions.
The domains $\mathcal{X}^{(j)} \subseteq \mathbb{R}^p$ are sets of vectors constructed by selecting from the inputs $\vc{x}$ only the $p$ coordinates $i^{(j)} \in \mathbb{N}_{mp}$ that correspond to the past of a single scalar time series $j$.
\begin{equation*}
\mathcal{X}^{(j)} = \{ \vc{x}^{(j)} : x^{(j)}_i = x_{i^{(j)}} \, \forall i, \ \vc{x} \in \mathcal{X}\}, \quad \cup_j \mathcal{X}^{(j)} = \mathcal{X} 
\end{equation*}

Further, instead of choosing the individual matrix-valued functions $\vc{H}^{(j)}$, we  propose to learn them.
We construct each $\vc{H}^{(j)}$ again as a sum of kernels
$\vc{H}^{(j)} = \sum_i^{s_j} \vc{H}^{(ji)}$ of possibly uneven number of summands $s_j$ of matrix-valued kernels $\vc{H}^{(ji)} : \mathcal{X}^{(j)} \times \mathcal{X}^{(j)} \to \mathbb{R}^{m \times m}$.
For this lowest level $\vc{H}^{(ji)}$ we focus on the family of decomposable kernels
$\vc{H}^{(ji)} = k^{(ji)} \, \vc{L}^{(ji)}$.
Here, the input kernels $k^{(ji)} : \mathcal{X}^{(j)} \times \mathcal{X}^{(j)} \to \mathbb{R}$ capturing the similarity between the inputs are fixed in advance from a dictionary of valid scalar-valued kernels (e.g. Gaussian kernels with varying scales).
The set $\mathcal{L} = \big\{\vc{L}^{(ji)} : j=\mathbb{N}_m, \, i=\mathbb{N}_{s_j}, \sum_j^m s_j = l \big\}$ of output kernels $\vc{L}^{(ji)} : \mathcal{Y} \to \mathcal{Y}$ encoding the relations between the outputs is learned within the cone of symmetric positive semidefinite matrices $\mathbb{S}^m_+$.
\begin{equation}
\vc{H} = \sum_{j=1}^m \vc{H}^{(j)} = \sum_{j=1}^m \sum_{i=1}^{s_j} \vc{H}^{(ji)} = \sum_{j=1}^m \sum_{i=1}^{s_j} k^{(ji)} \, \vc{L}^{(ji)}
\end{equation}

\subsection{Kernel Learning and Function Estimation}\label{sec:JointLearn}

Learning all the output kernels $\vc{L}^{(ji)}$ as full PSD matrices implies learning more than $m^3$ parameters.
To improve the generalization capability, we reduce the complexity of the problem drastically by restricting the search space for $\vc{L}$'s to PSD diagonal matrices  $\mathbb{D}^m_+$.
This essentially corresponds to the assumption of no contemporaneous relationships between the series.
We return to this point in Sect. \ref{sec:Comments}.

As explained in Sect. \ref{sec:Granger}, Granger (non-)causality learning typically searches for sparse models.
We bring this into our methods by imposing a further sparsity inducing regularizer $\mathit{Q}: (\mathbb{R}^{m \times m})^l \to \mathbb{R}$ on the set of the output kernels $\mathcal{L}$.
We motivate and elaborate suitable forms of $\mathit{Q}$ in Sect. \ref{sec:Sparsity}.

The joint learning of the kernels and the function can now be formulated as the problem of finding the minimising solution $\vc{f} \in \mathcal{F}$ and $\vc{L}$'s $\in \mathbb{D}^m_+$ of the regularised functional
\begin{equation}\label{eq:JointMin}
\mathit{J}(\vc{f},\mathcal{L}) := \mathit{R}(\vc{f}) + \tau \mathit{Q}(\mathcal{L}), \quad \tau > 0 \enspace ,
\end{equation}
where $\mathit{R}(\vc{f})$ is the regularised risk from \eqref{eq:GenMin}.
By calling on the properties of the RKHS, we reformulate this as a finite dimensional problem that can be addressed by conventional finite-dimensional optimisation approaches.
We introduce the gram matrices $\vc{K}^{(ji)} \in \mathbb{S}^n_+$ such that $K^{(ji)}_{ts} = k^{(ji)}(\vc{x}^{(j)}_t,\vc{x}^{(j)}_s)$ for all $t,s \in \mathbb{N}_n$, the output data matrix $\vc{Y} \in \mathbb{R}^{n \times m}$ such that $\vc{Y} = (\vc{y}_1, \ldots \vc{y}_n)^T$, and the coefficient matrix $\vc{C} \in \mathbb{R}^{n \times m}$ such that $\vc{C} = (\vc{c}_1, \ldots \vc{c}_n)^T$.

Using these and \eqref{eq:ReprTheor} it is easy to show that the minimisation of the regularised risk $\mathit{R}(\vc{f})$ in \eqref{eq:GenMin} with respect to $\vc{f} \in \mathcal{F}$ is equivalent to the minimisation with respect to $\vc{C} \in \mathbb{R}^{n \times m}$ of the
% finite dimensional
objective
\begin{equation}
\widetilde{R}(\vc{C}) := || \vc{Y} - \sum_{ji} \vc{K}^{(ji)} \vc{C} \vc{L}^{(ji)} ||_F^2 +
\lambda \sum_{ji} \langle  \vc{C}^T \vc{K}^{(ji)} \vc{C}, \vc{L}^{(ji)} \rangle_F  \enspace .
\end{equation}
The finite dimensional equivalent of \eqref{eq:JointMin} is thus the joint minimisation of 
\begin{equation}\label{eq:JointFin}
\widetilde{J}(\vc{C},\mathcal{L}) := \widetilde{R}(\vc{C},\mathcal{L}) + \tau \mathit{Q}(\mathcal{L}) \enspace .
\end{equation}

\subsection{Sparse Regularization}\label{sec:Sparsity}

The construction of the kernel $\vc{H}$ and the function space $\mathcal{F}$ described in Sect. \ref{sec:FuncSpace} imposes on the function $\vc{f}$ the necessary structure that allows the Granger-causality analysis (as per our 2nd objective set-out in Sect. \ref{sec:ProblemFormulation}). As explained in Sect. \ref{sec:Granger}, the other ingredient we need to identify the Granger non-causalities is sparsity within the structure of the learned function.

In our methods, the sparsity is introduced by the regularizer $\mathit{Q}$.
By construction of the function space, we can examine the elements of the output kernels $\vc{L}^{(ij)}$ (their diagonals) to 
make statements about the Granger non-causality.
We say the $j$-th scalar time series is non-causal for the $s$ series (given all the remaining series in the process) if $\vc{L}^{(ji)}_{ss} = 0$ for all $i \in \mathbb{N}_{s_j}$.

Essentially, any of the numerous regularizers that exist for sparse or structured sparse learning \cite{Bach2012a} could be used as $Q$, possibly based on some prior knowledge about the underlying dependencies within the time-series process.

We elaborate here two cases that do not assume any special structure in the dependencies as the base scenarios. 
The first
is the entry-wise $\ell_1$ norm across all the output kernels so that 
\begin{equation}\label{eq:L1}
\mathit{Q}_1(\mathcal{L}) = \sum_{ji} ||\vc{L}^{(ji)}||_1 = \sum_{ji} \sum_s^m |L^{(ji)}_{ss}| \enspace .
\end{equation}
The second
is the  $\ell_1/ \ell_2$ grouped norm 
\begin{equation}\label{eq:L12}
\mathit{Q}_{1/2}(\mathcal{L}) = \sum_{js} \sqrt{\sum_{i} \big(L^{(ji)}_{ss} \big)^2} \enspace .
\end{equation}
After developing the learning strategy for these in Sect(s). \ref{sec:GroupLasso} and \ref{sec:GroupMKL}, we provide some more intuition of their effects on the models and link to some other known graphical Granger techniques in Sect. \ref{sec:Comments}.

\section{Learning Strategy}\label{sec:LearnStrat}

First of all, we simplify the final formulation of the problem \eqref{eq:JointFin} in Sect. \ref{sec:JointLearn}.
Rather than working with a set of diagonal matrices $\vc{L}^{(ji)}$, we merge the diagonals into a single matrix $\vc{A}$.
We then re-formulate the problem with respect to this single matrix in place of the set and show how this reformulation can be suitably decomposed into smaller independent sub-problems.

We develop fit-to-purpose approaches for our two regularisers in Sect(s). \ref{sec:GroupLasso} and \ref{sec:GroupMKL}.
The first - based on the decomposition of the kernel matrices into the corresponding empirical features and on the variational formulation of norms \cite{Bach2012a} - shows the equivalence of the problem with group lasso \cite{Yuan2006,Zhao2006}.
The second proposes a simple alternating minimisation algorithm to obtain the two sets of parameters.

We introduce the non-negative matrix $\vc{A} \in \mathbb{R}^{l \times m}_+$ such that 
\begin{equation}
\vc{A} = \big( diag(\vc{L}^{11}), \ldots, diag(\vc{L}^{m s_m}) \big)^T
\end{equation}
(each row in $\vc{A}$ corresponds to  the diagonal of one output kernel; if $s_j = 1$ for all $j$ we have $A_{js} = \vc{L}^{(j1)}_{ss}$).
Using this change of variable, the optimisation problem \eqref{eq:JointFin} can be written equivalently as 
\begin{eqnarray}\label{eq:JointGamma}
& \argmin_{\vc{A}, \vc{C}} \, {\ddot{J}}(\vc{C},\vc{A}) & \nonumber \\
& \ddot{J}(\vc{C},\vc{A}) := \ddot{R}(\vc{C},\vc{A}) + \tau \ddot{Q}(\vc{A}) &  \enspace ,
\end{eqnarray}
where 
\begin{eqnarray}\label{eq:RGamma}
\ddot{R}(\vc{C},\vc{A}) & =  &
\sum_s^m \Big( || \vc{Y}_{:s} - \sum_{ji}  A_{(ji)s} \vc{K}^{ij} \vc{C}_{:s} ||_2^2 
 + \, \lambda \, \sum_{ji} A_{(ji)s} \vc{C}_{:s}^T \vc{K}^{(ji)} \vc{C}_{:s} \Big)   \nonumber \\
& = & \sum_s^m \Big(  \ddot{R}_s (\vc{C}_{:s},\vc{A}_{:s}) \Big) \enspace ,
\end{eqnarray}
and $\ddot{Q}(\vc{A})$ is the equivalent of $Q(\mathcal{L})$ so that
\begin{equation}\label{eq:GL1}
\ddot{Q}_1(\vc{A}) = ||\vc{A}||_1 = \sum_{rs} |A_{rs}|
\end{equation}
and 
\begin{equation}\label{eq:GL12}
\ddot{Q}_{1/2}(\vc{A}) = \sum_{js} \sqrt{\sum_i \big( A_{(ji)s} \big)^2}
\end{equation}
In equations \eqref{eq:RGamma} and \eqref{eq:GL12} we somewhat abuse the notation by using $\sum_{ji}  A_{(ji)s}$ to indicate the sum across the rows of the matrix $\vc{A}$.

From \eqref{eq:RGamma}-\eqref{eq:GL12} we observe that, with both of our regularizers, problem \eqref{eq:JointGamma} is conveniently separable along $s$ into the sum of $m$ smaller independent problems, one per scalar output series.
These can be efficiently solved in parallel, which makes our method scalable to very large multivariate systems.
The final complexity depends on the choice of the regulariser $\mathit{Q}$ and the appropriate algorithm.
The overhead cost can be significantly reduced by precalculating the gram matrices $\vc{K}^{(ij)}$ in a single preprocessing step and sharing these in between the $m$ parallel tasks.

\subsection{Learning with $\ell_1$ Norm}\label{sec:GroupLasso}

To unclutter notation we replace the bracketed double superscripts $(ij)$ by a single superscript $d = 1,\ldots,l$.
We also drop the regularization parameter $\tau$ (fix it to $\tau = 1$) as it is easy to show that any other value can be absorbed into the rescaling of $\lambda$ and the $\vc{C}$ and $\vc{A}$ matrices.
For each of the $s$ parallel tasks we indicate $\vc{A}_{:s} = \vc{a}$, $\vc{C}_{:s} = \vc{c}$ and $\vc{Y}_{:s} = \vc{y}$
so that the individual problems are the minimisations with respect to $\vc{a} \in \mathbb{R}^l_+$ and $\vc{c} \in \mathbb{R}^n$ of
\begin{equation}\label{eq:SingleProb}
\mathit{P}(\vc{c},\vc{a}) := || \vc{y} - \sum_d  a_d \vc{K}^d \vc{c} ||_2^2
+ \lambda \, \sum_{ji} a_d \vc{c}^T \vc{K}^d \vc{c} + \sum_d a_d  \enspace .
\end{equation}

We decompose (for example by eigendecomposition) each of the gram matrices as $\vc{K}^d = \vc{\Phi}^d (\vc{\Phi}^{d})^T$, where $\vc{\Phi}^d \in \mathbb{R}^{n \times n}$ is the matrix of the empirical features, and we introduce the variables $\vc{z}^d = a_d (\vc{\Phi}^{d})^T \vc{c} \in \mathbb{R}^n$ and the set $\mathcal{Z} = \{\vc{z}^d : \vc{z}^d \in \mathbb{R}^n, \, d \in \mathbb{N}_l\}$.
Using these we rewrite\footnote{We extend the function $x^2/y : \mathbb{R} \times \mathbb{R}_{+} \to \mathbb{R}_+$ to the point $(0 , 0)$ by taking the convention $0/0 = 0$.} equation \eqref{eq:SingleProb}
\begin{equation}\label{eq:SingleProbGZ}
\widetilde{P}(\mathcal{Z},\vc{a}) := 
|| \vc{y} - \sum_d  \vc{\Phi}^{d} \vc{z}^d ||_2^2
+ \sum_d \left( \frac{\lambda ||\vc{z}^d||_2^2}{a_{d}} + a_d \right) \enspace .
\end{equation}

We first find the closed form of the minimising solution for $\vc{a}$ as $a_d = \sqrt{\lambda} ||\vc{z}^d||_2$ for all $d$. Plugging this back to \eqref{eq:SingleProbGZ} we obtain 
\begin{equation}\label{eq:SingleProbZ}
\min_{\vc{a}} \widetilde{P}(\mathcal{Z},\vc{a}) = || \vc{y} - \sum_d  \vc{\Phi}^{d} \vc{z}^d ||_2^2
+ 2 \sqrt{\lambda} \sum_d ||\vc{z}^d||_2 \enspace .
\end{equation}

Seen as a minimisation with respect to the set $\mathcal{Z}$ this is the classical group-lasso formulation with the empirical features $\vc{\Phi}^d$ as inputs.
Accordingly, it can be solved by any standard method for group-lasso problems such as the proximal gradient descent method, e.g. \cite{Bach2012a}, which we employ in our experiments.
After solving for $\mathcal{Z}$ we can directly recover $\vc{a}$ from the above minimising identity and then obtain the parameters $\vc{c}$ from the set of linear equations
\begin{equation}\label{eq:SolveC}
(\sum_d a_{d} \vc{K}^d + \lambda \vc{I}_n) \, \vc{c} = \vc{y} \enspace .
\end{equation}

The algorithm outlined above takes advantage of the convex group-lasso reformulation \eqref{eq:SingleProbZ} and has the standard convergence and complexity properties of proximal gradient descent.
The empirical features $\vc{\Phi}^d$ can be pre-calculated and shared amongst the $m$ tasks to reduce the overhead cost.

\subsection{Learning with $\ell_1/\ell_2$ Norm}\label{sec:GroupMKL}

For the $\ell_1/\ell_2$ regularization, we need to return to the double indexation $(ji)$ to make clear how the groups are created.
As above, for each of the $s$ parallel tasks we use the vectors $\vc{a}, \vc{c}$ and $\vc{y}$.
However, for vector $\vc{a}$ we will keep the $(ji)$ notation for its elements. 
The individual problems are the minimisations with respect to $\vc{a} \in \mathbb{R}^l_+$ and $\vc{c} \in \mathbb{R}^n$ of
\begin{equation}\label{eq:SingleProbL12}
\mathit{P}(\vc{c},\vc{a}) :=  || \vc{y} - \sum_{ji}  a_{(ji)} \vc{K}^{(ji)} \vc{c} ||_2^2  
 + \lambda \, \sum_{ji} a_{(ji)} \vc{c}^T \vc{K}^{(ji)} \vc{c} + \sum_j \sqrt{ \sum_i a_{(ji)}^2 } 
\end{equation}

We propose to use the alternating minimisation with a proximal gradient step.
At each iteration, we alternatively solve for $\vc{c}$ and $\vc{a}$.
For fixed $\vc{a}$ we obtain $\vc{c}$ from the set of linear equations \eqref{eq:SolveC}.
With fixed $\vc{c}$, problem \eqref{eq:SingleProbL12} is a group lasso for $\vc{a}$ with groups defined by the sub-index $j$ within the double $(ji)$ indexation of the elements of $\vc{a}$. 
Here, the proximal gradient step takes place to move along the descend direction for $\vc{a}$. 
Though convex in $\vc{a}$ and $\vc{c}$ individually, the problem \eqref{eq:SingleProbL12} is jointly non-convex and therefore can converge to local minima.

\section{Interpretation and Crossovers}\label{sec:Comments}

To help the understanding of the inner workings of our methods and especially the effects of the two regularizers, we discuss here the crossovers to other existing methods for MKL and Granger modelling.

\paragraph{$\ell_1$ Norm} The link to group-lasso demonstrated in Sect. \ref{sec:GroupLasso} is not in itself too surprising. 
The formulation in \eqref{eq:SingleProb} can be recognised as a sparse multiple kernel learning problem which has been previously shown to relate to group-lasso (eg. \cite{Bach2008}, \cite{Xu2010}).
We derive this link in Sect. \ref{sec:GroupLasso} using the empirical feature representation to i) provide better intuition for the structure of the learned function $\vc{\widehat{f}}$,
ii) develop an efficient algorithm for solving problem \eqref{eq:SingleProb}.

The re-formulation 
in terms of the empirical features $\vc{\Phi}^d$ creates an intuitive bridge to the classical linear models.
Each $\vc{\Phi}^d$ can be seen as a matrix of features generated from a subset $\mathcal{X}^{(j)}$ of the input coordinates relating to the past of a single scalar time series $j$.
The group-lasso regularizer in equation \eqref{eq:SingleProbZ} has a sparsifying effect at the level of these subsets zeroing out (or not) the whole groups of parameters $\vc{z}^d$.
In the context of linear methods, this approach is known as the grouped graphical Granger modelling \cite{Lozano2009}.

Within the non-linear approaches to time series modelling, Sindhwani et al. \cite{Sindhwani2014} recently derived a similar formulation.
There the authors followed a strategy of multiple kernel learning from a dictionary of input kernels combined with a single learned output kernel (as opposed to our multiple output kernels).
They obtain their IKL model, which is in its final formulation equivalent to problem \eqref{eq:SingleProb}, by fixing the output kernel to identity. 

Though we initially formulate our problem quite differently, the diagonal constraint we impose on the output kernels essentially prevents the modelling of any contemporaneous relationships between the series (as does the identity output kernel matrix in IKL).
What remains in our methods are the diagonal elements, which are non-constant and sparse, and which can be interpreted as the weights of the input kernels in the standard MKL setting. 

\paragraph{$\ell_1/\ell_2$ Norm} The more complex $\ell_1/\ell_2$ regularisation discussed in Sect. \ref{sec:GroupMKL} is to the best of our knowledge novel in the context of multiple kernel learning.
It has again a strong motivation and clear interpretation in terms of the graphical Granger modelling.
The norm has a sparsifying effect not only at the level of the individual kernels but at the level of the groups of kernels operating over the same input partitions $\mathcal{X}^{(j)}$.
In this respect our move from the $\ell_1$ to the $\ell_1/\ell_2$ norm has a parallel in the same move in linear graphical Granger techniques.
%in the use of the same norms on the parameters of linear model.
The $\ell_1$ norm Lasso-Granger method \cite{Arnold2007} imposes the sparsity on the individual elements of the parameter matrices in a linear model, while the $\ell_1/\ell_2$ of the grouped-Lasso-Granger \cite{Lozano2009} works with groups of the corresponding parameters of a single input series across the multiple lags $p$.

\section{Experiments}\label{sec:Experiments}

To document the performance of our method, we have conducted a set of experiments on real and synthetic datasets. 
In these we simulate real-life forecasting exercise by splitting the data into a training and a hold-out set which is unseen by the algorithm when learning the function $\vc{\widehat{f}}$ and is only used for the final performance evaluation.

We compare our methods with the output kernel $\ell_1$ regularization (NVARL1) and and $\ell_1/\ell_1$ (NVARL12) with simple baselines (which nevertheless are often hard to beat in practical time series forecasting) as well as with the state-of-the-art techniques for forecasting and Granger modelling.
Namely, we compare with simple mean and univariate linear autoregressive models (LAR), multivariate linear vector autoregressive model with $\ell_2$ penalty (LVARL2), the group-lasso Granger method \cite{Lozano2009} (LVARL1), and a sparse MKL without the $\mathcal{X}^{(j)}$ input partitioning (NVAR).
Of these, the last two are the most relevant competitors. LVARL1, similarly to our methods, aims at recovering the Granger structure but is strongly constrained to linear modelling only.
NVAR has no capability to capture the Granger relationships but, due to the lack of structural constraints, it is the most flexible of all the models.

We evaluate our results with respect to the two objectives for the function $\vc{\widehat{f}}$ defined in Sect. \ref{sec:ProblemFormulation}.
We measure the accuracy of the one-step ahead forecasts by the mean square error (MSE) for the whole multivariate process averaged over 500 hold-out points.
The structural objective allowing the analysis of dependencies between the sub-processes is wired into the method itself (see Sect(s). \ref{sec:FuncSpace} and \ref{sec:Granger}) and is therefore satisfied by construction.
We produce adjacency matrices of the graphs of the learned dependencies, compare these with the ones produced by the linear Granger methods and comment on the observed results.

\subsection{Technical Considerations}
For each experiment we preprocessed the data by removing the training sample mean and rescaling with the training sample standard deviation.
We fix the number of kernels for each input partition to six ($s_j = 6$ for all $j$) and use the same kernel functions for all experiments: a linear, 2nd order and 3rd polynomial, and Gaussian kernels with width 0.5, 1 and 2.
We normalise the kernels so that the training Gram matrices have trace equal to the size of the training sample.

We search for the hyper-parameter $\lambda$ by a 5-fold cross-validation within a 15-long logarithmic grid $\lambda \in \{10^{-3},\ldots,10^{4} \}\sqrt{n}l$, where $n$ is the training sample size and $l$ is the number of kernels or groups (depending on the method).
In each grid search, we use the previous parameter values as warm starts.
We do not perform an exhaustive search for the optimal lag for each of the scalar input series by some of the classical testing procedures (based on AIC, BIC etc.).
We instead fix it to $p=5$ for all series in all experiments and rely on the regularization to control any excess complexity.

We implemented our own tools for all the tested methods based on variations of proximal gradient descent with ISTA line search \cite{Beck2009}.
The full Matlab code is available at \url{https://bitbucket.org/dmmlgeneva/nonlinear-granger}

\subsection{Synthetic Experiments}

{%% MG: this is just to allow that the matrix inline equation breaks at the end of line
    \def\OldComma{,}
    \catcode`\,=13
    \def,{%
      \ifmmode%
        \OldComma\discretionary{}{}{}%
      \else%
        \OldComma%
      \fi%
    }%
We have simulated data from a five dimensional non-Gaussian time-series process generated through a linear filter of a 5-dimensional i.i.d. exponential white noise $\vc{e}_t$ with identity covariance matrix (re-centered to zero and re-scaled to unit variance).
The matrix $\vc{\Psi} = [0.7,1.3,0,0,0; 0,0.6,-1.5,0,0; 0,-1.2,1.46,0,0; 0,0,0,0.6,1.4; 0,0,0,1.3,-0.5]$ in the filter $\vc{y}_t = \vc{e}_t + \vc{\Psi} \vc{e}_{t-1}$ is such that the process consists of two independent internally interrelated sub-processes, one composed of the first 3 scalar series, the other of the remaining two series.
This structural information, though known to us, is unknown to the learning methods (not considered in the learning process).
}

We list in Table \ref{tab:mseSynthetic} the predictive performance of the tested methods in terms of the average hold-out MSE based on training samples of varying size. 
Our methods clearly outperform all the linear models.
The functionally strongly constrained linear LVARL1 performs roughly on par with our methods for the small sample sizes.
But for larger sample sizes, the higher flexibility of the function space in our methods yields significantly more accurate forecasts (as much as 10\% MSE improvement).

\begin{table}
\setlength{\tabcolsep}{3pt} % General space between cols (6pt standard)
\caption{Synthetic experiments: MSE \tiny{(std)} \normalsize for 1-step ahead forecasts (hold-out sample average)}
\label{tab:mseSynthetic}
\begin{center}
\vskip -10pt
\begin{tabular}{l || c  c | c  c | c  c }
\hline
\hline
Train size&\multicolumn{2}{c|}{300} &\multicolumn{2}{c|}{700}& \multicolumn{2}{c|}{1000}\\
\hline
 Mean&0.925&\tiny{(0.047)}&0.923&\tiny{(0.047)}&0.923&\tiny{(0.047)}\\
 LAR&0.890&\tiny{(0.045)}&0.890&\tiny{(0.044)}&0.890&\tiny{(0.044)}\\
 LVAR&0.894&\tiny{(0.045)}&0.836&\tiny{(0.041)}&0.763&\tiny{(0.035)}\\
 LVARL1&0.787&\tiny{(0.037)}&0.737&\tiny{(0.031)}&0.722&\tiny{(0.030)}\\
 NVAR&0.835&\tiny{(0.041)}&0.735&\tiny{(0.032)}&0.719&\tiny{(0.030)}\\
\hdashline
 NVARL1&\textit{0.754}&\tiny{(0.034)}&0.706&\tiny{(0.030)}&\textbf{0.679}&\tiny{(0.028)}\\
 NVARL12&0.808&\tiny{(0.040)}&0.710&\tiny{(0.031)}&0.684&\tiny{(0.029)}\\
\hline \hline
Train size&\multicolumn{2}{c|}{1500} &\multicolumn{2}{c|}{2000}& \multicolumn{2}{c|}{3000}\\
\hline
 Mean&0.923&\tiny{(0.047)}&0.922&\tiny{(0.047)}&0.922&\tiny{(0.047)}\\
 LAR&0.888&\tiny{(0.045)}&0.889&\tiny{(0.045)}&0.888&\tiny{(0.045)}\\
 LVAR&0.751&\tiny{(0.034)}&0.741&\tiny{(0.033)}&0.687&\tiny{(0.028)}\\
 LVARL1&0.710&\tiny{(0.029)}&0.701&\tiny{(0.028)}&0.693&\tiny{(0.028)}\\
 NVAR&0.699&\tiny{(0.028)}&0.682&\tiny{(0.027)}&0.662&\tiny{(0.026)}\\
\hdashline
 NVARL1&\textit{\textbf{0.654}}&\tiny{(0.026)}&\textit{\textbf{0.640}}&\tiny{(0.025)}&\textbf{0.626}&\tiny{(0.025)}\\
 NVARL12&\textbf{0.659}&\tiny{(0.027)}&0.685&\tiny{(0.028)}&\textbf{0.657}&\tiny{(0.027)}\\

\hline
\hline
\end{tabular}
\end{center}
In brackets is the average standard deviation (std) of the MSEs. Results for NVARL1 and NVARL12 in \textbf{bold} are significantly better than \textbf{all} the linear competitors, in \textit{italics} are significantly better than the non-linear NVAR (using one-sided paired-sample t-test at 10\% significance level).
\end{table}

\begin{figure}
\centering
\begin{subfigure}{.5\textwidth}
  \centering
  \includegraphics[width=1\linewidth]{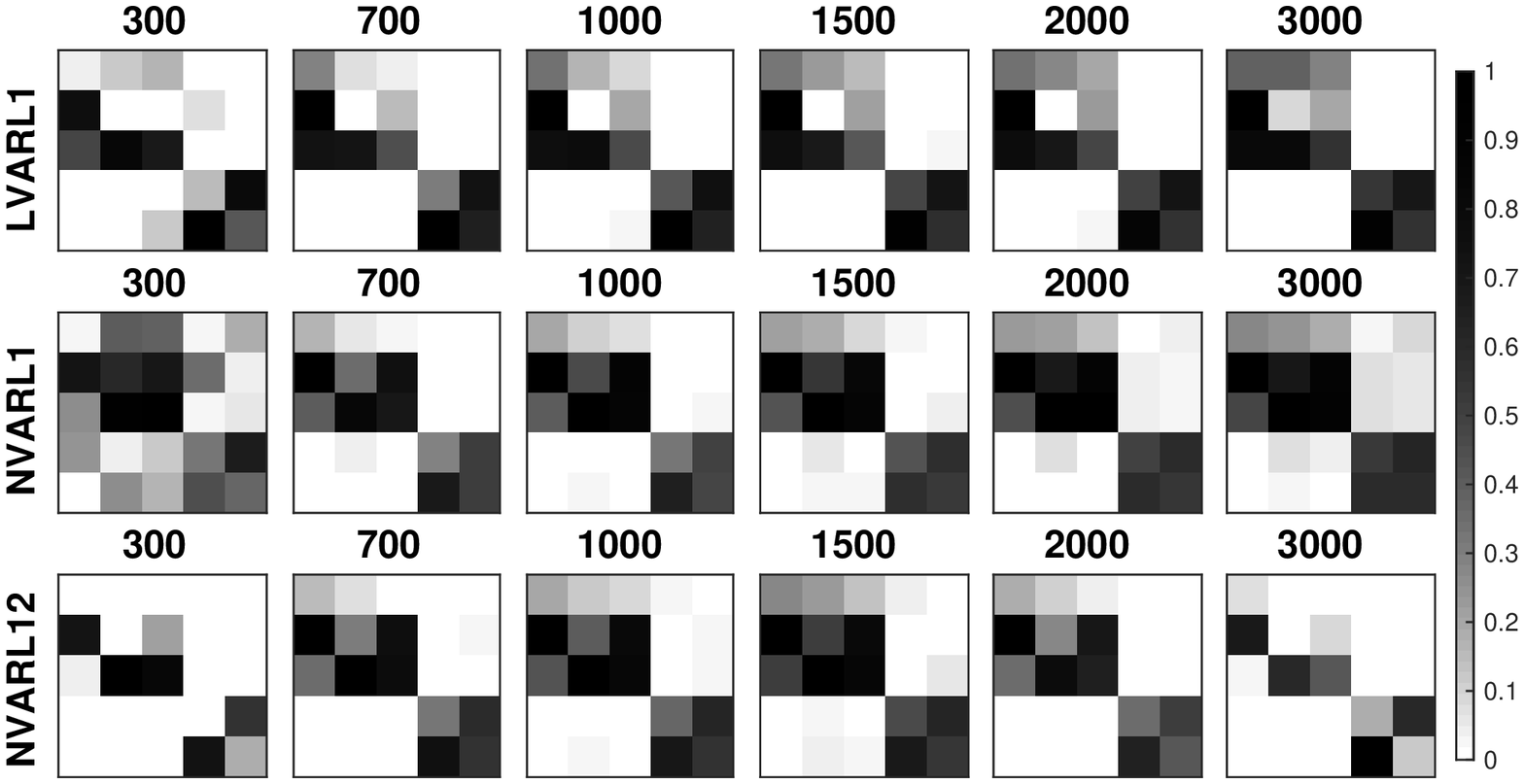}
  \caption{Synthetic}
  \label{fig:sub1}
\end{subfigure}%
\begin{subfigure}{.5\textwidth}
  \centering
  \includegraphics[width=1\linewidth]{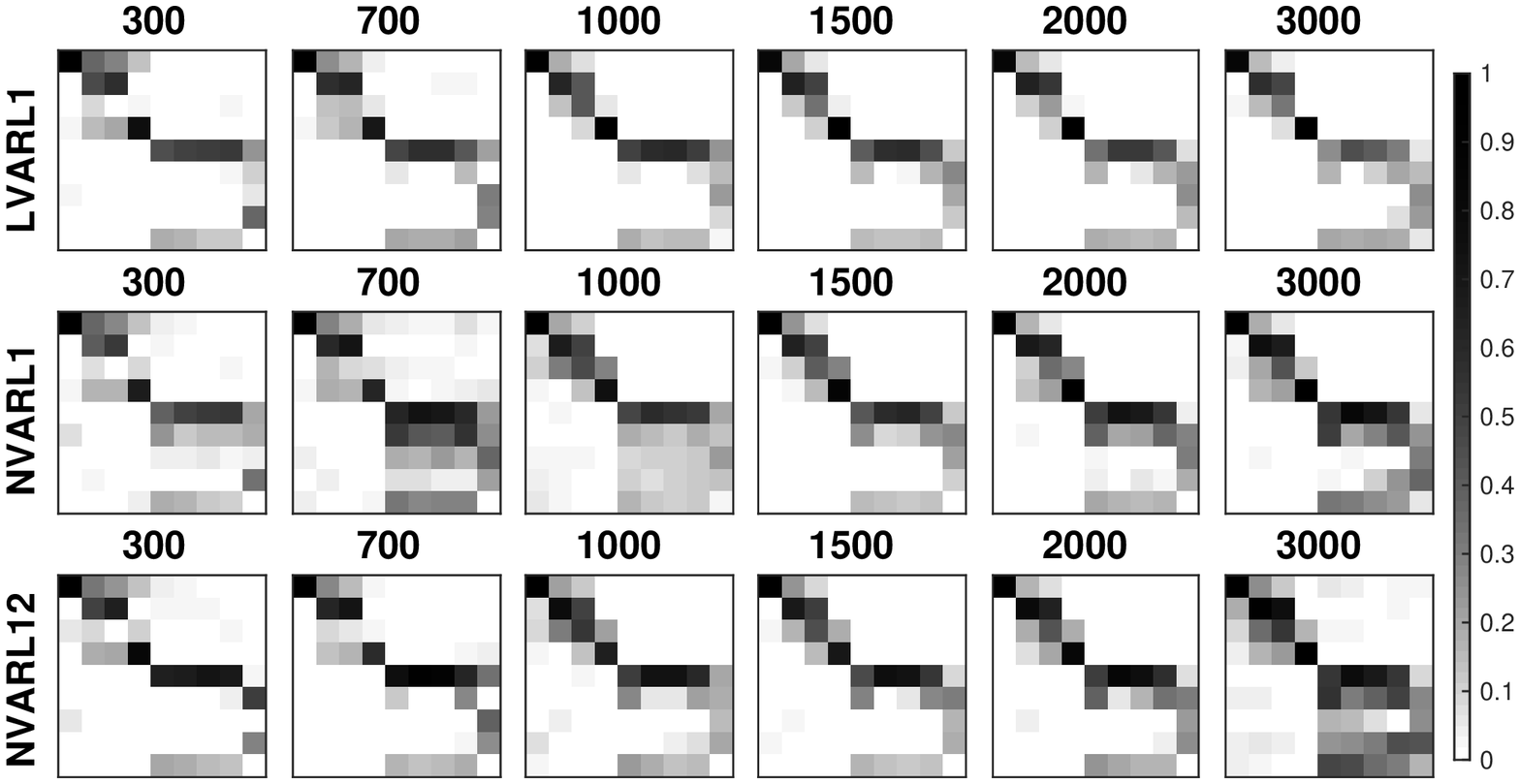}
  \caption{Real-data}
  \label{fig:sub2}
\end{subfigure}
\caption{Schematics of the learned adjecency matrices of Granger-causality graphs for the three sparse learning methods across varying training sample size. A scalar time series $y_i$ does not Granger-cause series $y_j$ (given all the other series) if the element $e_{ij}$ in the adjecency matrix is zero (white). The displayed adjecency matrices were derived from the learned matrices $\vc{A}$ by summing the respective elements across individual kernels. The values are rescaled so that the largest element in each matirx is equal to 1 (black).}\label{fig:granger}
\end{figure}

The structural constraints in our methods also help the performance when competing with the unstructured NVAR method, which has mostly less accurate forecasts.
At the same time, as illustrated in Fig. \ref{fig:granger}, our methods are able to correctly recover the Granger-causality structure (splitting the process into the two independent subprocesses by the zero off-diagonal blocks), which NVAR by construction cannot.

\subsection{Real Data Experiments}
We use data on water physical discharge publicly available from the website of the Water Services of the US geological survey (\url{http://www.usgs.gov/}).
Our dataset consists of 9 time series of daily rates of year-on-year growth at measurement sites along the streams of Connecticut and Columbia rivers.

The prediction accuracy of the tested methods is listed in Table \ref{tab:mseReal}.
Our non-linear methods perform on par with the state-of-the-art linear models.
This on one hand suggests that for the analysed dataset the linear modelling seems sufficient.
On the other hand, it confirms that our methods, which in general have the ability to learn more complex relationships by living in a richer functional space, are well behaved and can capture simpler dependencies as well. 
The structure encoded into our methods, however, benefits the learning since the unstructured NVAR tends to perform less accurately.

%\begin{small}
\begin{table}
\setlength{\tabcolsep}{3pt} % General space between cols (6pt standard)
\caption{Real-data experiments: MSE \tiny{(std)} \normalsize for 1-step ahead forecasts (hold-out sample average)}
\label{tab:mseReal}
\begin{center}
\vskip -10pt
\begin{tabular}{l || c  c | c  c | c  c }
\hline
\hline
Train size&\multicolumn{2}{c|}{300} &\multicolumn{2}{c|}{700}& \multicolumn{2}{c|}{1000}\\
\hline
 Mean&0.780&\tiny{(0.053)}&0.795&\tiny{(0.054)}&0.483&\tiny{(0.026)}\\
LAR&0.330&\tiny{(0.023)}&0.340&\tiny{(0.024)}&0.152&\tiny{(0.013)}\\
LVARL2&0.302&\tiny{(0.021)}&0.311&\tiny{(0.022)}&0.140&\tiny{(0.012)}\\
LVARL1&0.310&\tiny{(0.022)}&0.310&\tiny{(0.023)}&0.140&\tiny{(0.012)}\\
NVAR&0.328&\tiny{(0.023)}&0.316&\tiny{(0.023)}&0.148&\tiny{(0.012)}\\
\hdashline
NVARL1&0.308&\tiny{(0.023)}&0.317&\tiny{(0.024)}&0.140&\tiny{(0.012)}\\
NVARL12&0.321&\tiny{(0.023)}&0.322&\tiny{(0.024)}&0.141&\tiny{(0.012)}\\
\hline \hline
Train size&\multicolumn{2}{c|}{1500} &\multicolumn{2}{c|}{2000}& \multicolumn{2}{c|}{3000}\\
\hline
 Mean&0.504&\tiny{(0.03)}&0.464&\tiny{(0.027)}&0.475&\tiny{(0.017)}\\
LAR&0.181&\tiny{(0.015)}&0.179&\tiny{(0.013)}&0.187&\tiny{(0.008)}\\
LVARL2&0.167&\tiny{(0.014)}&0.164&\tiny{(0.013)}&0.170&\tiny{(0.007)}\\
LVARL1&0.165&\tiny{(0.014)}&0.163&\tiny{(0.013)}&0.170&\tiny{(0.008)}\\
NVAR&0.169&\tiny{(0.014)}&0.166&\tiny{(0.012)}&0.173&\tiny{(0.007)}\\
\hdashline
NVARL1&0.164&\tiny{(0.014)}&0.161&\tiny{(0.013)}&0.167&\tiny{(0.007)}\\
NVARL12&0.162&\tiny{(0.014)}&0.160&\tiny{(0.012)}&0.166&\tiny{(0.007)}\\

\hline
\hline
\end{tabular}
\end{center}
In brackets is the average standard deviation (std) of the MSEs.
\vskip -5pt
\end{table}
%\end{small}

The learned dynamical dependence structure of the time series is depicted in Fig. \ref{fig:granger}.
In the dataset (and the adjacency matrices), the first 4 series are the Connecticut measurement sites starting from the one highest up the stream and moving down to the mouth of the river.
The next 5 our the Columbia measurement sites ordered in the same manner.

From inspecting the learned adjacency matrices, we observe that all the sparse methods recover similar Granger-causal structures.
Since we do not know the ground truth in this case, we can only speculate about the accuracy of the structure recovery.
Nevertheless, it seems plausible that there is little dynamical cross-dependency between the Connecticut and Columbia measurements as the learned graphs suggest (the two rivers are at the East and West extremes of the US).

\section{Conclusions}
We have developed a new method for forecasting and Granger-causality modelling in multivariate time series that does not rely on prior assumptions about the shape of the dynamical dependencies (other than being sparse). 
The method is based on learning a combination of multiple operator-valued kernels in which the multiple output kernels are learned as sparse diagonal matrices.
We have documented on experiments that our method outperforms linear competitors in the presence of strong non-linearities and is able to correctly recover the Granger-causality structure which non-structured kernel methods cannot do.

\subsection*{Acknowledgements}
This work was partially supported by the research projects HSTS (ISNET) and RAWFIE \#645220 (H2020).
We thank Francesco Dinuzzo for helping to form the initial ideas behind this work through fruitful discussions while visiting in IBM Research, Dublin.

\newpage

\bibliographystyle{splncs03}
\bibliography{kernelTS_ECML2017}

\begin{thebibliography}{10}
\providecommand{\url}[1]{\texttt{#1}}
\providecommand{\urlprefix}{URL }

\bibitem{Arnold2007}
Arnold, A., Liu, Y., Abe, N.: {Temporal causal modeling with graphical granger
  methods}. Proceedings of the 13th ACM SIGKDD international conference on
  Knowledge discovery and data mining - KDD '07  (2007)

\bibitem{Bach2008}
Bach, F.: {Consistency of the group lasso and multiple kernel learning}. The
  Journal of Machine Learning Research  (2008)

\bibitem{Bach2012a}
Bach, F., Jenatton, R., Mairal, J., Obozinski, G.: {Optimization with
  sparsity-inducing penalties}. Foundations and Trends in Machine Learning
  (2012)

\bibitem{Bahadori2013}
Bahadori, M., Liu, Y.: {An Examination of Practical Granger Causality
  Inference}. SIAM Conference on Data Mining  (2013)

\bibitem{Beck2009}
Beck, A., Teboulle, M.: {Gradient-based algorithms with applications to signal
  recovery}. Convex Optimization in Signal Processing and Communications
  (2009)

\bibitem{Brockwell2006}
Brockwell, P.J., Davis, R.A.: {Time Series: Theory and Methods}. Springer
  Science+Business Media, LLC, 2nd edn. (2006)

\bibitem{Caponnetto2008}
Caponnetto, A., Micchelli, C.A., Pontil, M., Ying, Y.: {Universal Multi-Task
  Kernels}. Machine Larning Research  (2008)

\bibitem{Dinuzzo2011a}
Dinuzzo, F., Ong, C.: {Learning output kernels with block coordinate descent}.
  In: International Conference on Machine Learning (ICML) (2011)

\bibitem{Eichler2012}
Eichler, M.: {Graphical modelling of multivariate time series}. Probability
  Theory and Related Fields  (2012)

\bibitem{Franz2006}
Franz, M.O., Sch{\"{o}}lkopf, B.: {A unifying view of wiener and volterra
  theory and polynomial kernel regression.} Neural computation  (2006)

\bibitem{Granger1969}
Granger, C.W.J.: {Investigating Causal Relations by Econometric Models and
  Cross-spectral Methods}. Econometrica: Journal of the Econometric Society
  (1969)

\bibitem{Jawanpuria2015}
Jawanpuria, P., Lapin, M., Hein, M., Schiele, B.: {Efficient Output Kernel
  Learning for Multiple Tasks}. In: NIPS (2015)

\bibitem{Kadri2012}
Kadri, H., Rakotomamonjy, A., Bach, F., Preux, P.: {Multiple Operator-valued
  Kernel Learning}. In: NIPS (2012)

\bibitem{Lanckriet2004}
Lanckriet, G.G.R., Cristianini, N., Bartlett, P., Ghaoui, L.E., Jordan, M.I.:
  {Learning the kernel matrix with semidefinite programming}. Journal of
  Machine Learning Research  (2004)

\bibitem{Lim2014}
Lim, N., D'Alch{\'{e}}-Buc, F., Auliac, C., Michailidis, G.: {Operator-valued
  Kernel-based Vector Autoregressive Models for Network Inference}. Machine
  Learning  (2014)

\bibitem{Lozano2009}
Lozano, A.C., Abe, N., Liu, Y., Rosset, S.: {Grouped graphical Granger modeling
  for gene expression regulatory networks discovery.} Bioinformatics (Oxford,
  England)  (2009)

\bibitem{Micchelli2005b}
Micchelli, C.A., Pontil, M.: {On learning vector-valued functions.} Neural
  computation  (2005)

\bibitem{Pillonetto2014}
Pillonetto, G., Dinuzzo, F., Chen, T., {De Nicolao}, G., Ljung, L.: {Kernel
  methods in system identification, machine learning and function estimation: A
  survey}. Automatica  (2014)

\bibitem{Sindhwani2014}
Sindhwani, V., Minh, H.Q., Lozano, A.: {Scalable Matrix-valued Kernel Learning
  for High-dimensional Nonlinear Multivariate Regression and Granger
  Causality}. In: UAI (2013)

\bibitem{Turkman2014}
Turkman, K.F., Scotto, M.G., {de Zea Bermudez}, P.: {Non-Linear Time Series}.
  {Springer} (2014)

\bibitem{Xu2010}
Xu, Z., Jin, R., Yang, H., King, I., Lyu, M.R.: {Simple and efficient multiple
  kernel learning by group lasso}. International Conference on Machine Learning
  (ICML)  (2010)

\bibitem{Yuan2006}
Yuan, M., Lin, Y.: {Model selection and estimation in regression with grouped
  variables}. Journal of the Royal Statistical Society: Series B (Statistical
  Methodology)  (2006)

\bibitem{Zhao2006}
Zhao, P., Rocha, G.: {Grouped and hierarchical model selection through
  composite absolute penalties} (2006)

\end{thebibliography}

\end{document}